\documentclass[a4paper,twocolumn,10pt,fleqn]{article}
\usepackage{times}
\usepackage[top=20mm,left=15mm,right=15mm,bottom=27mm,nohead,nofoot]{geometry}

\usepackage{latexsym}
\usepackage{graphicx}
\usepackage{pdfpages,epstopdf}
\DeclareGraphicsExtensions{.pdf}
\usepackage{amsmath}  
\usepackage{amsfonts}
\usepackage{amssymb}
\usepackage{amsthm}
\usepackage[square,comma,numbers,sort&compress]{natbib}
\newenvironment{inmargins}[1]{\begin{list}{}{
  \leftmargin=#1 \rightmargin=#1 \parsep=0pt
  \partopsep=0pt}\item[]}{\end{list}}

\makeatletter
\def\section{\@startsection {section}{1}{0mm}
{-2.5ex plus -1ex minus -.2ex}{0.5ex plus 0.2ex}{\large\bf}}
\def\subsection{\@startsection{subsection}{2}{0mm}
{-2.0ex plus -1ex minus -.2ex}{0.5ex plus 0.2ex}{\bf}}
\def\subsubsection{\@startsection{subsubsection}{3}{0mm}
{-1.5ex plus -1ex minus -.2ex}{0.5ex plus 0.2ex}{\rm}}
\renewcommand \thesection {\@arabic\c@section.}
\renewcommand\thesubsection   {\@arabic\c@section.\@arabic\c@subsection}

\long\def\@makecaption#1#2{%
  \vskip\abovecaptionskip
  \sbox\@tempboxa{#1~ #2}%
  \ifdim \wd\@tempboxa >\hsize
    #1~ #2\par
  \else
    \global \@minipagefalse
    \hb@xt@\hsize{\hfil\box\@tempboxa\hfil}%
  \fi
  \vskip\belowcaptionskip}
\makeatother

\newtheorem{definition}{Definition}

\newcommand{\G}{\mathcal{G}}
\newcommand{\R}{\mathbb{R}}
\newcommand{\HQ}{\mathbb{H}}
\newcommand{\N}{\mathbb{N}}
\newcommand{\be}{\begin{equation}}
\newcommand{\ee}{\end{equation}}
\newcommand{\bv}[1]{\mathrm{\mathbf{#1}}}

\newcommand{\LC}{{\sf \hspace*{2ex}\rule{0.15ex}%
       {1.5ex}\hspace*{-1.5ex}\rule{1.5ex}{0.15ex}\hspace*{0.5ex}}}

\begin{document}

\twocolumn[
\vspace*{10pt}
\begin{center}
{\Large \bf Geometric operations implemented by conformal geometric algebra neural nodes
\\[6pt]}
{\large  Eckhard HITZER
\\[4pt]}
{\large 
University of Fukui%
}

\end{center}

\begin{inmargins}{15mm}
   {\bf Abstract: } 
Geometric algebra is an optimal frame work for calculating with vectors. The geometric algebra of a space includes elements that represent all the its subspaces (lines, planes, volumes, ...). Conformal geometric algebra expands this approach to elementary representations of arbitrary points, point pairs, lines, circles, planes and spheres. Apart from including curved objects, conformal geometric algebra has an elegant unified quaternion like representation for all proper and improper Euclidean transformations, including reflections at spheres, general screw transformations and scaling. Expanding the concepts of real and complex neurons we arrive at the new powerful concept of conformal geometric algebra neurons. These neurons can easily take the above mentioned geometric objects or sets of these objects as inputs and apply a wide range of geometric transformations via the geometric algebra valued weights.  
\end{inmargins}
\vspace*{20pt}
]

\section{Introduction}

The co-creator of calculus W. Leibniz dreamed of
a new type of mathematics in which every number, every operation
and every relation would have a clear geometric counterpart. 
Subsequently the inventor of the concept of vector space
and our modern notion of algebra H. Grassmann was officially
credited to fulfill Leibniz's vision. Contemporary to
Grassmann was W. Hamilton, who took great pride in establishing
the algebra of rotation generators in 3D, which he himself called
quaternion algebra. About 30 years later W. Clifford successfully fused
Grassmann's and Hamilton's work together in what he called 
geometric algebra. Geometric algebra can be understood as
an algebra of a vector space and all its subspaces equipped
with an associative and invertible geometric product of vectors. 

It is classic knowledge that adding an extra dimension for the
origin point to 
a vector space leads to the projective geometry of rays, where
Euclidean points correspond to rays. Adding one more dimension 
for the point of infinity allows to treat lines as circles through
infinity, and planes as spheres through infinity and unifies their
treatment. This socalled conformal geometric algebra represents
geometric points, spheres and planes by 5D vectors. The inner product
of two conformal points yields their Euclidean distance. Orthogonal
transformations preserve inner products and in the 5D model 
Euclidean distances, provided that they also keep the point at infinity
invariant. According to Cartan and Dieudonn\'{e} all orthogonal
transformations are products of reflections. Proper and improper Euclidean transformations can therefore
be expressed nowadays in conformal geometric algebra as elegant
as complex numbers express rotations in 2D. And the language is not 
limited to 3+2D, $n$+2D follows the very same principles. 

Transformation groups generated by products of reflections in
geometric algebra are known as Clifford (or Lipschitz or versor) groups~\cite{HLR:GS1,PL:Clifford}. Versors (Clifford group or Lipschitz elements) are simply
the geometric products of the normal vectors to the (hyper) planes
of reflection. These versors assume the role of geometric weights
in concept of conformal geometric algebra neural nodes. Precursors for these
nodes are complex~\cite{AH:book}, quaternion~\cite{NM:Matsui},
and Clifford spinor~\cite{BS:ESANN2000,SB:thesis} neurons. 
They have also been named versor Clifford neurons~\cite{BHT:CVN}, but regarding
their fundamental geometric nature even the term \textit{geometric neurons} seems fully justified.

\section{Geometric algebra}

\begin{definition}[Clifford geometric algebra]
\label{df:GA}
A Clifford geometric algebra $\G_{p,q}$ is defined by the
associative geometric product of elements of a quadratic vector space
$\R^{p,q}$, their linear combination and closure. $\G_{p,q}$ includes 
the field of real numbers $\R$ and the vector space $\R^{p,q}$ as
subspaces. The geometric product of two vectors is defined as
\be
  \label{eq:geopro}
  \bv{a}\bv{b} = \bv{a}\cdot \bv{b} +\bv{a}\wedge \bv{b},
\ee
where $\bv{a}\cdot \bv{b}$ indicates the standard inner product and 
the bivector $\bv{a}\wedge \bv{b}$ indicates Grassmann's antisymmetric
outer product. $\bv{a}\wedge \bv{b}$ can be geometrically interpreted as
the oriented parallelogram area spanned by the vectors $\bv{a}$ and $\bv{b}$.
Geometric algebras are graded, with grades (subspace dimensions) ranging from 
zero (scalars) to $n=p+q$ (pseudoscalars, $n$-volumes).
\end{definition}

For example geometric algebra $\G_{3}=\G_{3,0}$ of three-dimensional Euclidean space 
$\R^3=\R^{3,0}$ has an eight-dimensional basis of scalars (grade 0), vectors (grade 1),
bivectors (grade 2) and trivectors (grade 3). Trivectors in $\G_{3}$ 
are also referred to as oriented volumes or pseudoscalars.
Using an orthonormal basis $\{\bv{e}_1, \bv{e}_2, \bv{e}_3\}$ for $\R^3$ 
we can write the basis of $\G_{3}$ as
\be
  \label{eq:G3basis}
  \{1, \bv{e}_1, \bv{e}_2, \bv{e}_3,
    \bv{e}_2\bv{e}_3, \bv{e}_3\bv{e}_1, \bv{e}_1\bv{e}_2,  
    i=\bv{e}_1\bv{e}_2\bv{e}_3\}.
\ee
In \eqref{eq:G3basis} $i$ is the unit trivector, i.e. the oriented volume
of a unit cube. The even subalgebra $\G_{3}^+$ of 
$\G_{3}$ 
is isomorphic to the quaternions $\HQ$ of Hamilton. We therefore call 
elements of $\G_{3}^+$ rotors, because 
they  rotate all  elements of $\G_{3}$ . 
The role of complex (and quaternion) conjugation is naturally taken
by reversion ($\bv{a}_1,\bv{a}_2,\ldots, \bv{a}_s \in {\R^{p,q}}$, $s \in \N$)
\be
  \label{eq:rev}
  (\bv{a}_1\bv{a}_2\ldots \bv{a}_s)^{\sim}=\bv{a}_s\ldots \bv{a}_2\bv{a}_1.
\ee

The inverse of a non-null
vector $\bv{a}\in {\R^{p,q}}$ is
\be
  \label{eq:vinv}
  \bv{a}^{-1} = \frac{\bv{a}}{\bv{a}^2}.
\ee
A reflection at a hyperplane  normal  $\bv{a}$ 
is
\be
  \label{eq:ref}
  \bv{x}^{\prime} = - \bv{a}^{-1}\bv{x}\bv{a}.
\ee

A rotation by the angle $\theta$ in the plane of a unit bivector $\bv{i}$
can thus be given as the product $R=\bv{a}\bv{b}$ of two vectors 
$\bv{a}$, $\bv{b}$ from the $\bv{i}$-plane (i.e. geometrically as
a sequence of two reflections) with angle $\theta/2$,
\be
  \label{eq:rot}  
  {\R^{p,q}} \ni \bv{x} \rightarrow  R^{-1}\bv{x}R \in {\R^{p,q}},
\ee
where the vectors $\bv{a}$, $\bv{b}$ are in the plane of the unit bivector
$\bv{i} \in \G_{p,q}$ if and only if $\bv{a}\wedge \bv{i} = \bv{b}\wedge \bv{i} = 0$.
The rotor $R$ can also be expanded as
\be
  \label{eq:rotexp}  
  R=\bv{a}\bv{b}= |\bv{a}||\bv{b}|exp(\theta\bv{i}/2),
\ee
where $|\bv{a}|$, and $|\bv{b}|$ are the lengths of $\bv{a}$, $\bv{b}$. This
description corresponds exactly to using quaternions.

Blades of grade $k, \,0\leq k \leq n=p+q$ are the outer products of $k$ vectors $\bv{a}_l$
($1\leq l\leq k$) and directly represent the $k$-dimensional
vector subspaces $V$ spanned by
the set of vectors $\bv{a}_l$ $(1\leq l\leq k)$. This is also called the outer product null space (OPNS) representation. 
\be
  \label{eq:OPNS}
  \bv{x} \in V = \mathrm{span}[\bv{a}_1,\ldots\bv{a}_k]
  \,\,\Leftrightarrow  \,\,
  \bv{x}\wedge \bv{a}_1\wedge\bv{a}_2 \wedge\ldots \wedge\bv{a}_k = 0.
\ee
Extracting a certain grade part from the geometric product of two 
blades $A_k$ and $B_l$ has a deep geometric meaning.

One example is the grade
$l-k$ part (contraction~\cite{DFM:GACS}) of the geometric product $A_kB_l$, 
that represents the orthogonal
complement of a $k$-blade $A_k$ in an $l$-blade $B_l$, provided that $A_k$ is contained in $B_l$
\be
 \label{eq:lcontr}
  A_k \LC B_l = \langle A_k B_l\rangle_{l-k}.
\ee
Another important grade part of the geometric product of $A_k$ and $B_l$ is the maximum
grade $l+k$ part, also called the outer product part
\be
  \label{eq:outp}
  A_k \wedge B_l = \langle A_k B_l\rangle_{l+k}.
\ee
If $A_k \wedge B_l$ is non-zero it represents the union of the disjoint 
(except for the zero vector)
subspaces represented by $A_k$ and $B_l$. 

The \textit{dual} of
a multivector $A$ is defined by geometric division with the pseudoscalar $I=$
$A^{\ast} = A \,I^-1$, which maps $k$-blades into $(n-k)$-blades, where $n=p+q$.
Duality transforms inner products (contractions) to outer products and vice versa. 
The outer product null space representation (OPNS) of \eqref{eq:OPNS} is therefore
transformed by duality into the socalled inner product null space (IPNS) 
representation
\be
  x \wedge A = 0 \Longleftrightarrow x \cdot A^{\ast} = 0.
\ee

\subsection{Conformal geometric algebra}

Conformal geometric algebra embeds the geometric
algebra $\G_3$ of $\R^3$ in the geometric algebra of $\R^{4,1} = \R^{3+1,0+1}$
Given an orthonormal basis for $\R^{4,1}$
\be
  \label{eq:confbas}
  \{\bv{e}_1,\bv{e}_2,\bv{e}_3,\bv{e}_+,\bv{e}_- \}
\ee
with 
\be
  \bv{e}_1^2=\bv{e}_2^2=\bv{e}_3^2=\bv{e}_+^2=-\bv{e}_-^2=1,
\ee
we introduce a change of basis for the two additional dimensions
$\{\bv{e}_+,\bv{e}_-\}$ by
\be   
  \bv{e}_{0} = \frac{1}{2}(\bv{e}_- -\bv{e}_+),
  \qquad
  \bv{e}_{\infty} = \bv{e}_- +\bv{e}_+.
\ee
The vectors $\bv{e}_{0}$ and $\bv{e}_{\infty}$ are \textit{isotropic} vectors, i.e.
\be   
  \bv{e}_{0}^2 = \bv{e}_{\infty}^2 = 0,
\ee
and have inner and outer products of
\be   
  \label{eq:prode0ei}
  \bv{e}_{0}\cdot \bv{e}_{\infty}=-1, 
  \qquad
  E = \bv{e}_{\infty} \wedge \bv{e}_{0} = \bv{e}_{+} \wedge \bv{e}_{-}.
\ee
We further have the following useful relationships
\begin{align}
  \label{eq:confrel}
  &\bv{e}_{0}E = -\bv{e}_{0}, 
  \,\,\,
  E\bv{e}_{0}=\bv{e}_{0},
  \bv{e}_{\infty}E = \bv{e}_{\infty}, 
 \nonumber \\
  &E \bv{e}_{\infty} = - \bv{e}_{\infty},
  \,\,\,
  E^2 = 1.
\end{align}

As we will now see, conformal geometric
algebra is advantageous for the unified
representation of many types of geometric 
transformations. In the next section we
will further consider the unified
representation of eight different
types of Euclidean geometric objects possible in 
conformal geometric algebra.

Combining reflections \eqref{eq:ref} leads to an overall sign (parity)  
for odd and even numbers of (reflection plane) vectors $\bv{a}$, $\bv{b}$, 
etc. Therefore we define the grade involution
\be
  \label{eq:grinv}
  \widehat{A} = \sum_{k=0}^{n} (-1)^k \langle A \rangle_k.
\ee
A {Clifford (or Lipschitz) group} is a subgroup in $\G_{p,q}$ generated by non-null vectors
$\bv{x}\in \R^{p,q}$ 
\be
  \label{eq:Cliffgr}
  {\Gamma_{p,q} = \{m \in \G_{p,q} 
             \mid \forall \bv{x}\in \R^{p,q}, \,\widehat{m}^{-1}\bv{x} \,m \,\in \R^{p,q}\}}
\ee
For every $m\in \Gamma_{p,q}$ we have $m \widetilde{m} \in \R$. 
Examples are $m=\bv{a}$ and $m=R=\bv{a}\bv{b}$ for reflections
and rotations, respectively.

Clifford groups include Pin($p,q$), Spin($p,q$), 
and $\mathrm{Spin}_+(p,q)$ groups as covering groups 
of {orthogonal $O(p,q)$}, special orthogonal $SO(p,q)$ and $SO_+(p,q)$ groups, respectively. 
{Conformal} transformation groups $C(p,q)$ preserve inner products (angles) 
of vectors in $\R^{p,q}$ up to a change of scale. 
$C(p,q)$ is isomorphic to $O(p+1,q+1)$. 

{The metric affine group (orthogonal transformations and \textit{translations})} of 
$\R^{p,q}$ is a subgroup  of $O(p+1,q+1)$,
and can be implemented as a {Clifford group} in $\G_{p+1,q+1}$.

\subsection{Geometric objects \label{sc:ConfObj}}

\begin{figure}[h]
\centering
 \resizebox{0.3\textwidth}{!}{\includegraphics{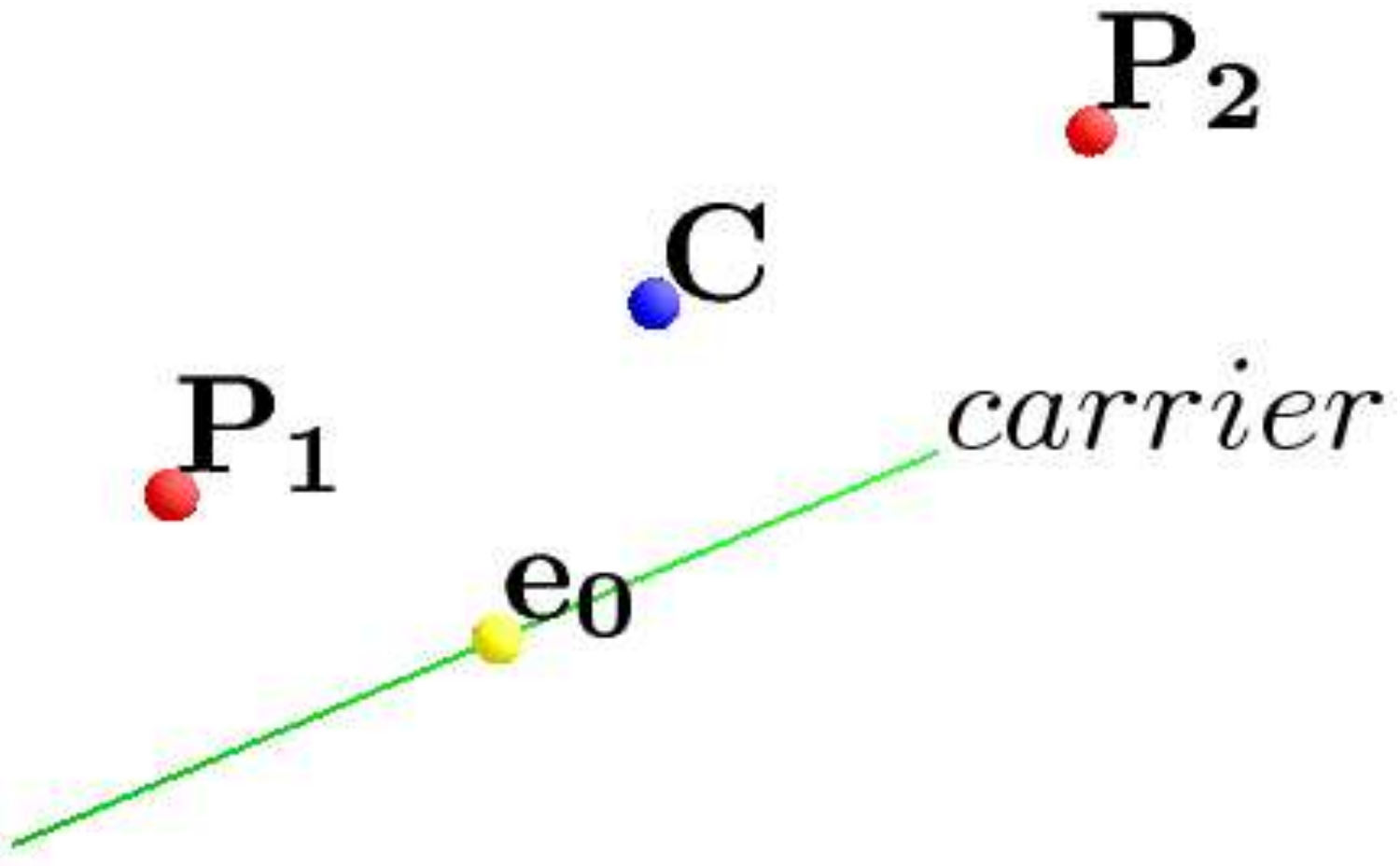}}
 \caption{Conformal point pair $P_1\wedge P_2$ with center $C$.}
 \label{fg:PointPair}
\end{figure}

The conformal geometric
algebra $\G_{4,1}$ provides us with a superb model~\cite{HLR:GS1,DFM:GACS,PA:ConfGr} 
of Euclidean geometry.  
The basic geometric objects in conformal geometric algebra are 
homogeneous conformal points
\be
  P=\bv{p}+\frac{1}{2}p^2\bv{e}_{\infty}+\bv{e}_{0}, 
  \label{eq:confP}
\ee
where $\bv{p} \in \R^3, p=\sqrt{\bv{p}^2}$.
The $+\bv{e}_{0}$ term shows that we include projective geometry. 
The second term $+\frac{1}{2}p^2\bv{e}_{\infty}$ ensures, that conformal
points are isotropic vectors
\be
  P^2 = PP = P\cdot P = 0. 
\ee
In general the inner product of two conformal points $P_1$ and $P_2$ 
gives their Euclidean distance
\be
  P_1 \cdot P_2 = -\frac{1}{2}(\bv{p}_1 - \bv{p}_2)^2.
\ee
Orthogonal transformations preserve this distance. 
The outer product of two conformal points
spans a conformal point pair (in OPNS) as in Fig. \ref{fg:PointPair}
\be
   Pp
   = P_1\wedge P_2 .
   \label{eq:pp1Dc}
\ee
This and the following illustrations were produced with the OpenSource
visual software CLUCalc~\cite{CP:CLUCalc}, which is also based on conformal geometric algebra. 

\begin{figure}[h]
\centering
 \resizebox{0.3\textwidth}{!}{\includegraphics{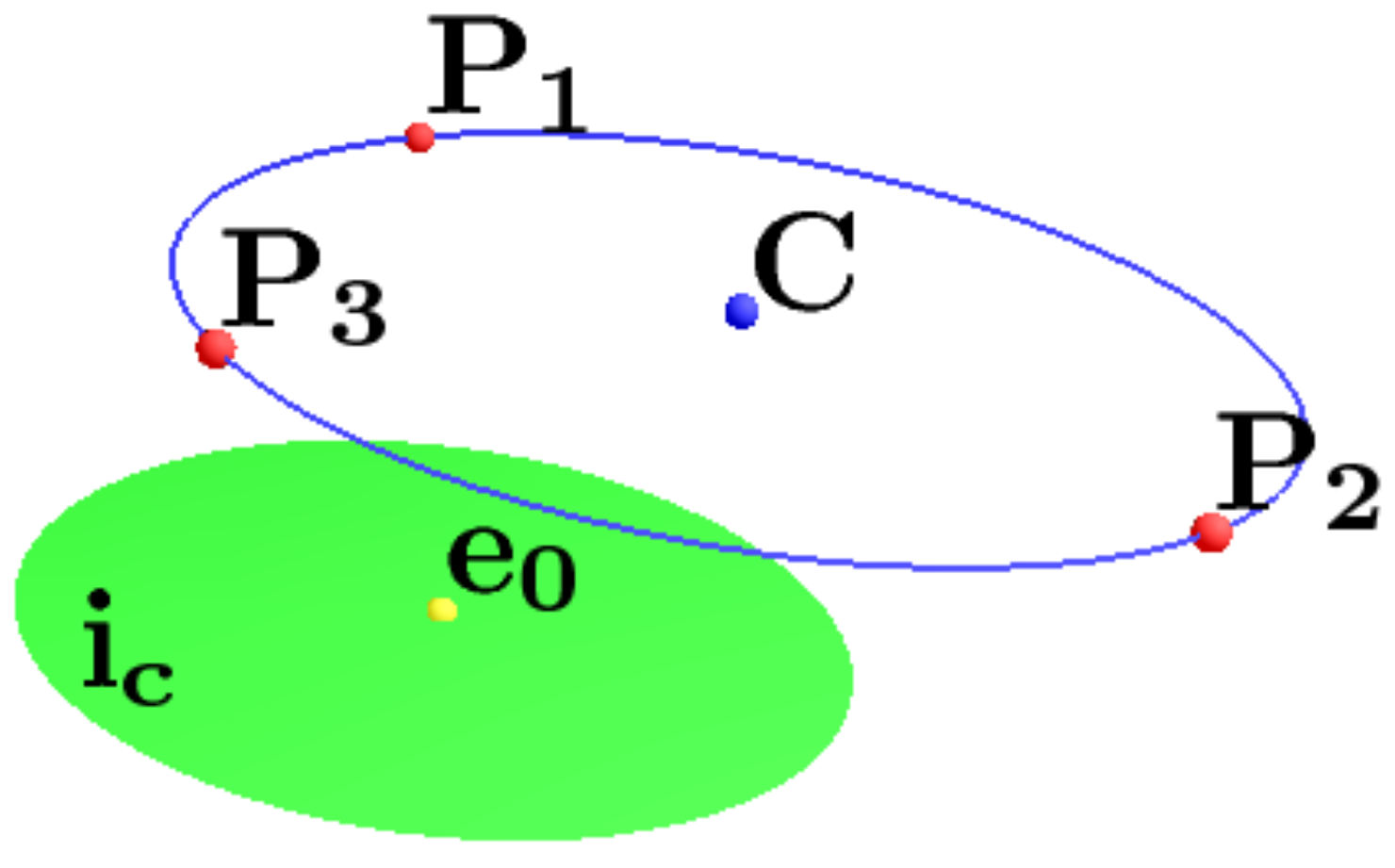}}
 \caption{Conformal circle $P_1\wedge P_2 \wedge P_3$ with center $C$.}
 \label{fg:Circle}
\end{figure}

The outer product three conformal points gives a circle (cf. Fig. \ref{fg:Circle})
\be
  \label{eq:circle}
  Circle 
    = P_1\wedge P_2 \wedge P_3.
\ee

\begin{figure}[h]
\centering
 \resizebox{0.3\textwidth}{!}{\includegraphics{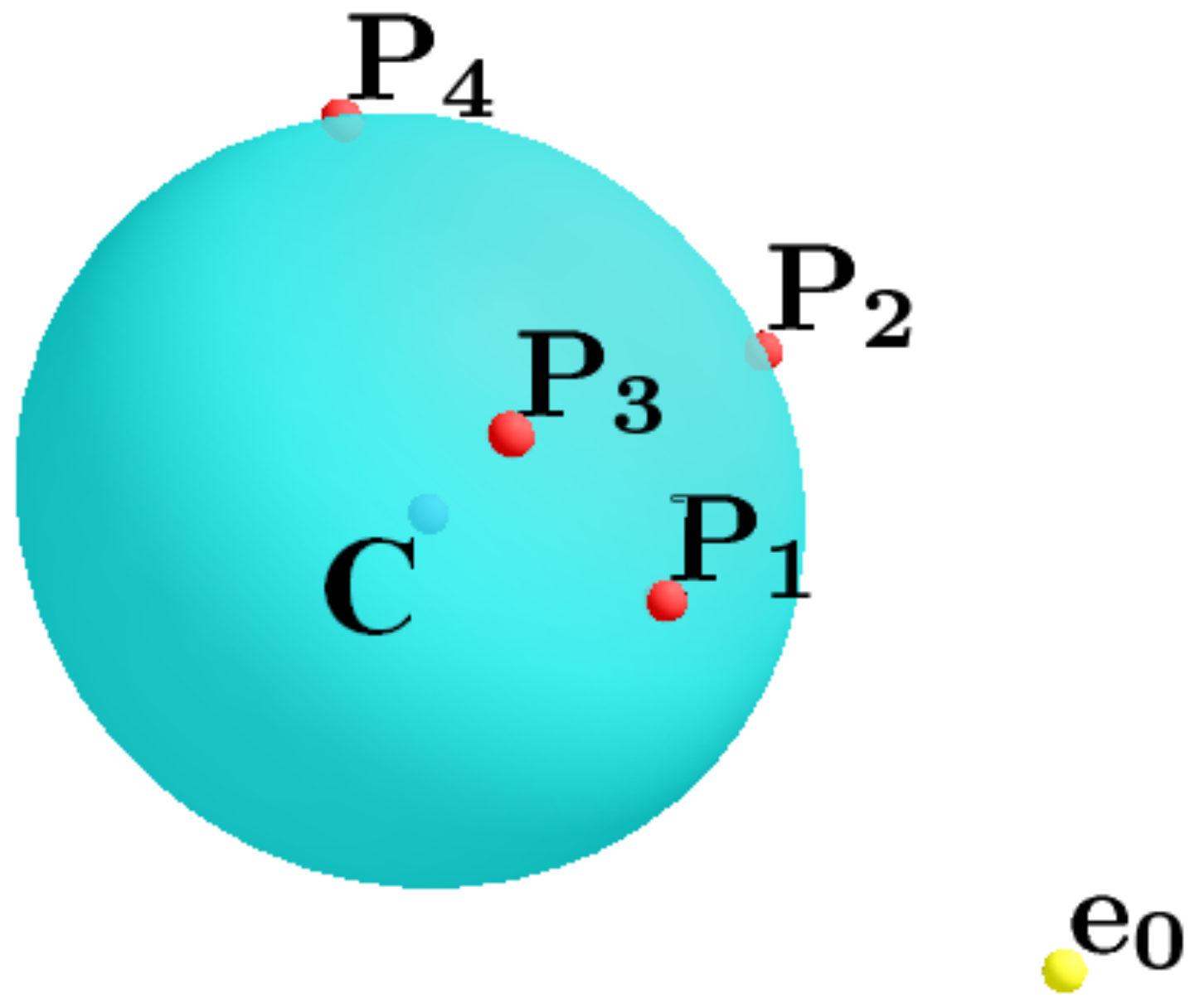}}
 \caption{Conformal sphere $P_1\wedge P_2 \wedge P_3 \wedge P_4$ with center $C$.}
 \label{fg:Sphere}
\end{figure}

The conformal outer product of four conformal points gives a sphere (cf. Fig. \ref{fg:Sphere})
\begin{eqnarray}
  Sphere
      &=& P_1\wedge P_2 \wedge P_3 \wedge P_4.
  \label{eq:V4conf}
\end{eqnarray}

\begin{figure}[h]
\centering
 \resizebox{0.3\textwidth}{!}{\includegraphics{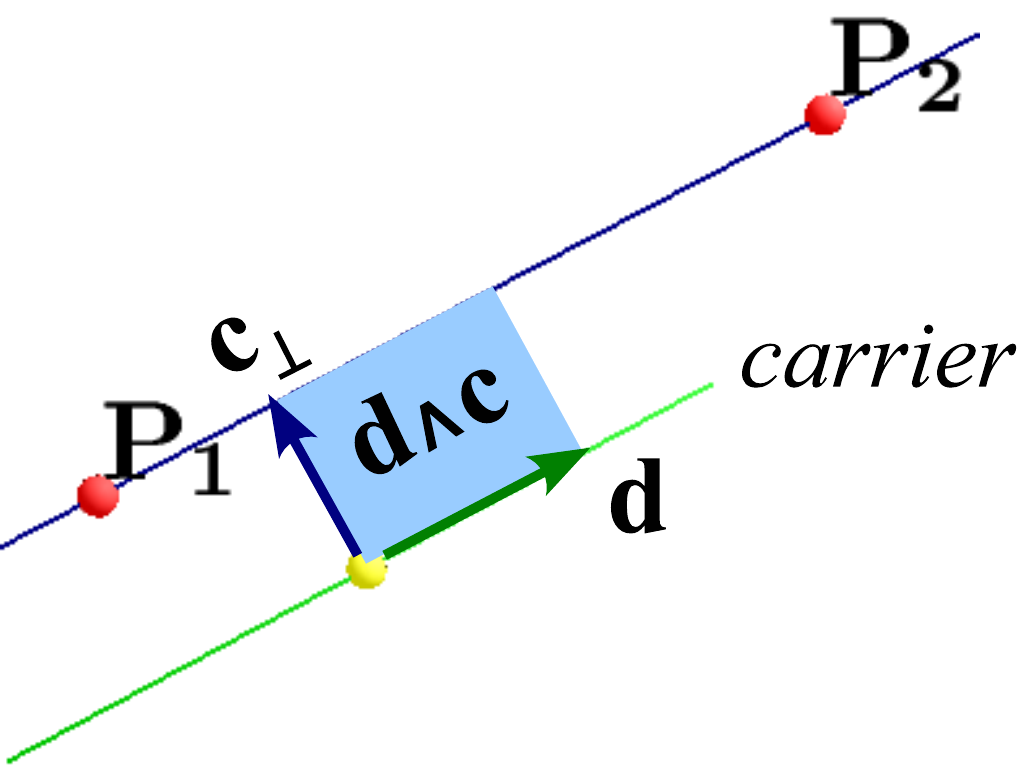}}
 \caption{Conformal line $P_1\wedge P_2 \wedge \bv{e}_{\infty}$ with 3D direction vector $\bv{d}$.}
 \label{fg:Line}
\end{figure}

\begin{figure}[h]
\centering
 \resizebox{0.3\textwidth}{!}{\includegraphics{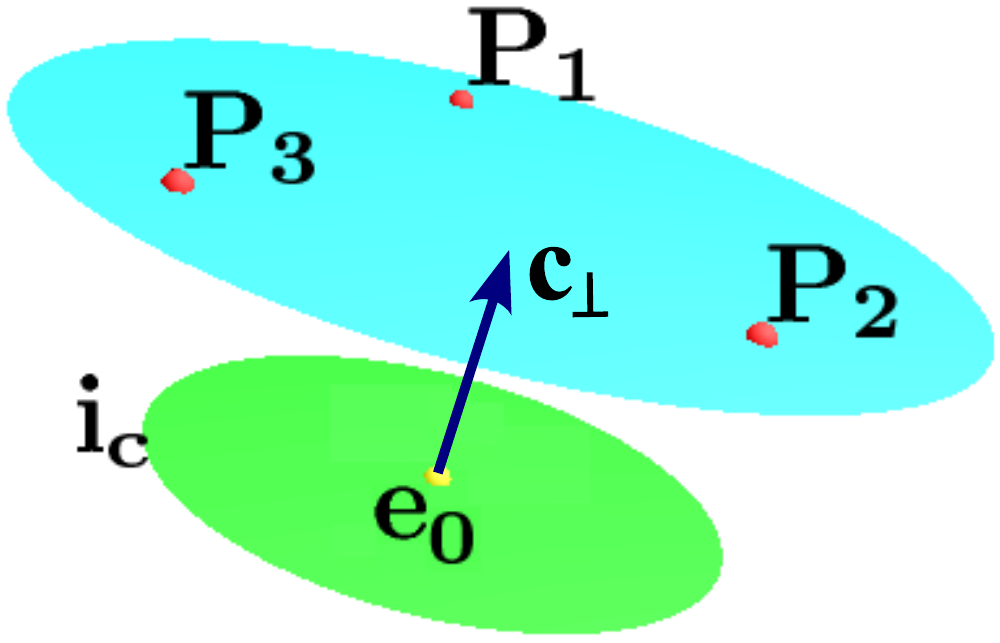}}
 \caption{Conformal plane $P_1\wedge P_2 \wedge P_3 \wedge \bv{e}_{\infty}$.}
 \label{fg:Plane}
\end{figure}

If one of the
points is at infinity, we get conformal lines (circles through infinity, cf. Fig. \ref{fg:Line})
\begin{align}
 Line 
   &= P_1 \wedge P_2 \wedge \bv{e}_{\infty}
   = Pp \wedge \bv{e}_{\infty}
   \nonumber \\
   &= 2r(\bv{d}\wedge \bv{c} \bv{e}_{\infty} - \bv{d} E )
   \propto \bv{m}\,\bv{e}_{\infty} - \bv{d} E,
   \label{eq:line1Dc}
\end{align}
where $\bv{d}$ is the direction vector of the line and $\bv{c}$ the 3D midpoint between $P_1$ 
and $P_2$, $2r=|\bv{p}_2-\bv{p}_2|$. $\bv{c}$ can be replaced by any point $\bv{p}$ on the line. 
$\bv{m}=\bv{d}\wedge \bv{c}=\bv{d}\wedge \bv{p}$
is also called bivector moment of the line.
Conformal planes (spheres through infinity, cf. Fig. \ref{fg:Plane}) are represented by
\be
  \label{eq:plane}
  Plane
    = P_1 \wedge P_2 \wedge P_3 \wedge \bv{e}_{\infty}
    = Circle \wedge \bv{e}_{\infty}.
\ee
Further flattened objects are
\be
  P \wedge \bv{e}_{\infty},
\ee
a (flat) finite--infinite point pair, and
\be
  -i_s E = Sphere \wedge \bv{e}_{\infty} \sim I_5=iE, 
\ee
the in 5D embedded (flat) {3D Euclidean space $\R^3$}.

\subsection{General reflection and motion operators (motors)}

As in $\G_{3}$, 3D rotations around the origin in the conformal model 
$\G_{4,1}$ are still represented by rotors \eqref{eq:rot}. 
The standard IPNS representation
of a conformal plane, i.e. the dual of the direct OPNS representation 
\eqref{eq:plane} results in a vector, that both describes the unit normal direction
$\bv{n} = \bv{c}_{\perp}/|\bv{c}_{\perp}|$ 
and the position (signed distance $d=\bv{c}_{\perp}\cdot\bv{n}$ 
from the origin) of a plane by
\be
  \label{eq:splane}
  \mu = Plane^{\ast} = \bv{n} + d \bv{e}_{\infty}. 
\ee
We can reflect the conformal point $X$ at the general plane \eqref{eq:splane} 
similar to \eqref{eq:ref}
with
\be
  \label{eq:planeref}
  X^{\prime} = -\mu^{-1} X \mu. 
\ee
A reflection of a general conformal object $O$ (point, point pair, line, ...,
sphere) at the general plane \eqref{eq:splane} 
similar to \eqref{eq:planeref}
with
\be
  \label{eq:planegref}
  O^{\prime} = \mu^{-1} \widehat{O} \mu,
\ee
where the grade involution takes care of the sign changes. 
Just as we obtained rotations \eqref{eq:rot} by double reflections 
we now obtain rotations around arbitrary axis (lines of intersection 
of two planes) by double reflections at planes $\mu_1$ and $\mu_2$
\be
  \label{eq:grot}
  X^{\prime} = R^{-1} X R, \qquad R = \mu_1 \mu_2.
\ee
And if the two planes $\mu_1$ and $\mu_2$ happen to be parallel, i.e.
$\bv{n}_1 = \pm \bv{n}_2$, we instead get a translation by twice the 
Euclidean distance $\bv{t}/2$ between the planes
\begin{align}
  \label{eq:trans}
  &X^{\prime} = T^{-1} X T, 
  \nonumber \\
  &T(\bv{t}) = \mu_1 \mu_2 
    = 1+\frac{1}{2}\bv{t} \bv{e}_{\infty} 
    = \exp(\frac{1}{2}\bv{t} \bv{e}_{\infty}).
\end{align}
A general motion operator (motor) results from combining rotations
\eqref{eq:grot} and translations \eqref{eq:trans} to
\be
  M = TR. 
\ee
The standard IPNS representation of a sphere results in the vector
\be
  \label{eq:SphereIPNS}
  \sigma = S^{\ast} = C -\frac{1}{2}r^2\bv{e}_{\infty},
\ee
which is exactly the dual $S^{\ast} = S \,i_3^{-1}E$ of \eqref{eq:V4conf}. 
The expression \eqref{eq:SphereIPNS} shows that in conformal GA a point can be regarded
as a sphere with zero radius. 
We can reflect (invert) the conformal point $X$ at the sphere \eqref{eq:SphereIPNS} 
[or at a conformal point]
similar to \eqref{eq:ref} and \eqref{eq:planeref} simply by
\be
  \label{eq:sphereref}
  X^{\prime} = -\sigma^{-1} X \sigma. 
\ee
The double reflection at two concentric spheres $\sigma_1$ and $\sigma_2$
(centered at $\bv{c}$) results
in a rescaling operation~\cite{DFM:GACS} with factor $s$ and center $\bv{c}$
\begin{align}
  \label{eq:scalor}
  &X^{\prime} = Z^{-1} X Z, 
  \nonumber \\
  &Z = \sigma_1 \sigma_2 = T^{-1}(\bv{c})\exp{(E\, \frac{1}{2}\log s)}\,T(\bv{c}).
\end{align}

Table \ref{tb:refls} summarizes the various reflections possible
in conformal GA. 
\begin{table}[h]
 \caption{Summary of reflections $O \rightarrow - W^{-1} \widehat{O} W$ in conformal GA. $O$ is a general
conformal object (point, point pair, circle, sphere, flat point, line, plane, 3D space). The conformal versor $W$
represents the mirror.}
 \label{tb:refls}
 \centering
 \begin{tabular}{|l|l|}
 \hline 
\rule[-1mm]{0mm}{4.0mm}%
   Mirror &  Versor $W$ \\ 
 \hline
  plane  & $\bv{n} + d \bv{e}_{\infty}$ \\ 
\rule[-1.5mm]{0mm}{4.9mm}%
  point  & $\bv{p}+\frac{1}{2}p^2\bv{e}_{\infty}+\bv{e}_{0}$ \\
\rule[-1.5mm]{0mm}{4.9mm}%
  sphere & $C -\frac{1}{2}r^2\bv{e}_{\infty}$ \\
  line   & $\bv{m}\,\bv{e}_{\infty} - \bv{d} E$ \\
 \hline
 \end{tabular}
\end{table}
Table \ref{tb:mots} summarizes the motions and scaling possible
in conformal GA. 
\begin{table}[h]
 \caption{Summary of motions and scaling $O \rightarrow W^{-1} {O} W$ in conformal GA. $O$ is a general
conformal object (point, point pair, circle, sphere, flat point, line, plane, 3D space). The conformal versor $W$
represents the motion operator.}
 \label{tb:mots}
 \centering
 \begin{tabular}{|l|l|}
 \hline
\rule[-1mm]{0mm}{4.0mm}%
   Motion operator &  Versor $W$ \\ 
 \hline
\rule[-1.5mm]{0mm}{4.9mm}%
  Rotor  & $exp(\theta\bv{i}/2)$ \\ 
\rule[-1.5mm]{0mm}{4.9mm}%
  Translator  & $\exp(\frac{1}{2}\bv{t} \bv{e}_{\infty})$ \\
\rule[-1.5mm]{0mm}{4.9mm}%
  Motor & $TR$ \\ 
\rule[-2mm]{0mm}{4.9mm}%
  Scale operator   & $T^{-1}(\bv{c})\exp{(E\, \frac{1}{2}\log s)}\,T(\bv{c})$ \\
 \hline
 \end{tabular}
\end{table}

\subsection{Combining elementary transformations}

\begin{figure}[h]
\centering
 \resizebox{0.4\textwidth}{!}{\includegraphics{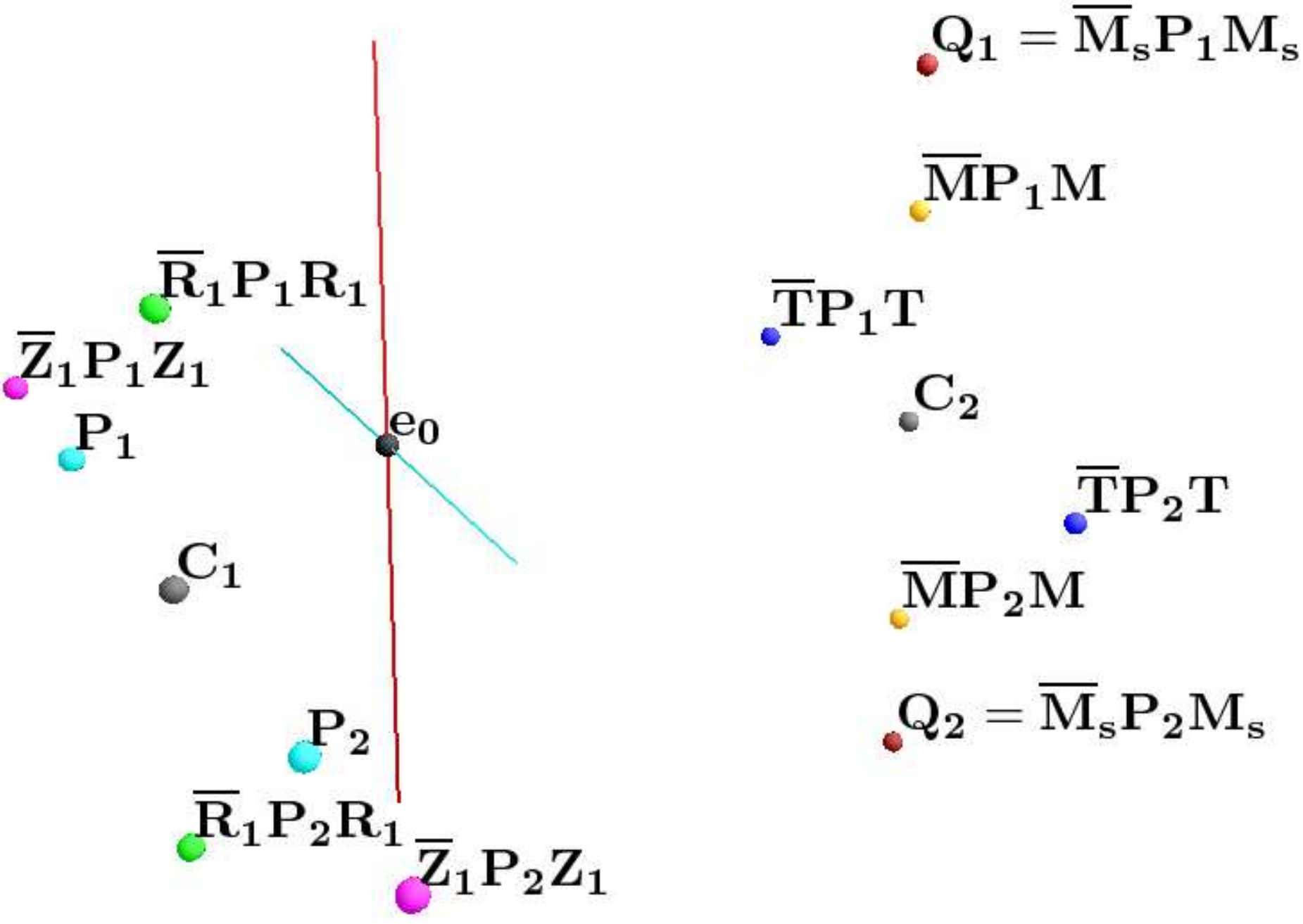}}
 \caption{Examples of combinations of geometric transformations: rotor $R$,
translator $T$, scaling operator $Z$, motor and scaling op. $M_s$.}
 \label{fg:PptoQq}
\end{figure}

All operations in Tables \ref{tb:refls} and \ref{tb:mots} can be combined to
give further geometric transformations, like rotoinversions, glide reflections, etc.!
Algebraically the combination is represented by simply computing the product of the
multivector versors of the elementary transformations explained above. 
Figure \ref{fg:PptoQq} shows some examples of combinations of geometric transformations
applied to a point pair $P_1 \wedge P_2$.
The overbars abbreviate the inverse of an operator. 

The following illustrations have been produced with the socalled
Space Group Visualizer~\cite{HP:SGV}, a symmetry visualization program for 
crystallographic space group, which is also based on conformal
geometric algebra software. 

\begin{figure}[h]
\centering
 \resizebox{0.4\textwidth}{!}{\includegraphics{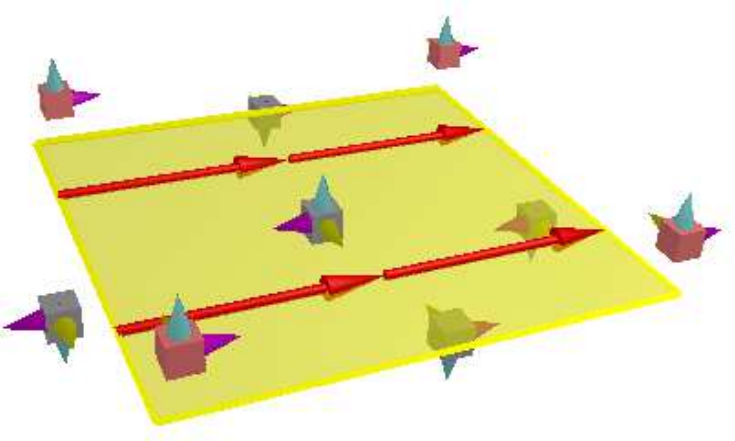}}
 \caption{Examples of glide reflection, which is a reflection combined
with a translation parallel to the plane of reflection.}
 \label{fg:glide}
\end{figure}

\begin{figure}[h]
\centering
 \resizebox{0.3\textwidth}{!}{\includegraphics{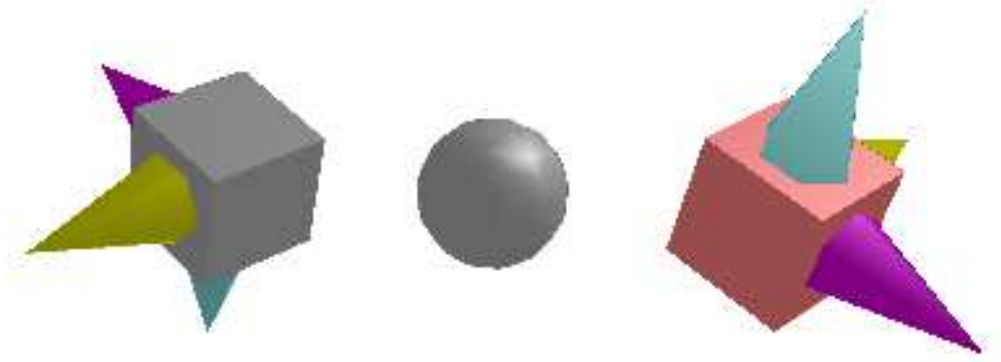}}
 \caption{Examples of point inversion.}
 \label{fg:inversion}
\end{figure}

\begin{figure}[h]
\centering
 \resizebox{0.4\textwidth}{!}{\includegraphics{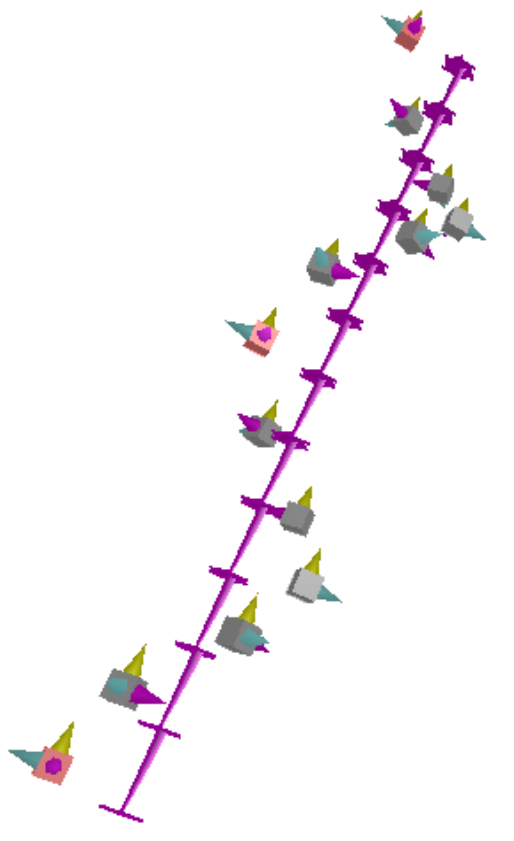}}
 \caption{Examples of a screw transformation.}
 \label{fg:screw}
\end{figure}

\begin{figure}[h]
\centering
 \resizebox{0.3\textwidth}{!}{\includegraphics{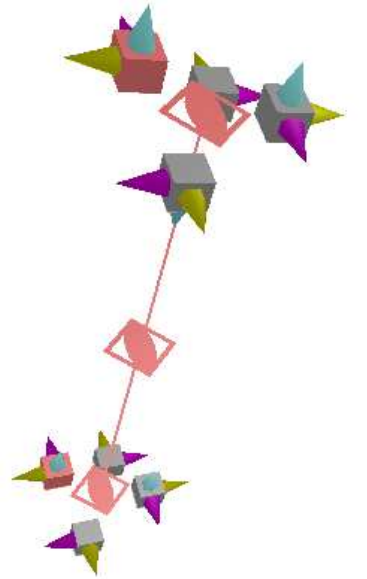}}
 \caption{Examples of a 90 degree rotoinversion, which is a combination of point inversion
and a 90 degree rotation.}
 \label{fg:rotoinversion}
\end{figure}

Figure \ref{fg:glide} shows a glide reflection, i.e. a reflection combined
with a translation parallel to the plane of reflection.
Figure \ref{fg:inversion} shows a point inversion. 
Figure \ref{fg:screw} shows a screw transformation, which is a rotation 
followed by a translation along the axis of the rotation. 
Figure \ref{fg:rotoinversion} shows a rotoinversion, which is a combination of point inversion
and rotation. A rotoinversion is also equivalent to a rotation followed by a reflection at a plane perpendicular to the axis of rotation.

\section{Geometric neurons}

Conformal versors $V$
describe in conformal Clifford group ~\cite{PL:Clifford} representations  
the above mentioned transformations 
of arbitrary conformal geometric object multivectors $X\in \mathcal{G}(\mathbb{R}^{3+1,1})$
of section \ref{sc:ConfObj}
\begin{align}
  \label{eq:vagain}
  X^{\prime} = (-1)^v V^{-1} X V,
\end{align}
where the versor
is a geometric product of $v$ \textit{invertible} 
vectors $\in \mathbb{R}^{3+1,1}$. 

The geometric neuron (GN) is characterized by a two-sided multiplication
of a single multivector weight versor $W \in \mathcal{G}(\mathbb{R}^{3+1,1})$,
input multivectors $X \in \mathcal{G}(\mathbb{R}^{3+1,1})$, 
and multivector thresholds $\Theta \in
\mathcal{G}(\mathbb{R}^{3+1,1})$ 
\begin{align}
  \label{eq:CVN}
  Y = (-1)^w W^{-1}X{W} + \Theta,
\end{align}
where $w$ represents the number of vector factors (parity) in $W$.
The theory, optimization and example simulation of such geometric neurons
has been studied in \cite{BHT:CVN}. It was shown that e.g. the inversion
at a sphere can be learned exactly by a geometric neuron, outperforming
linear networks and multilayer perceptrons with the same or higher 
number of degrees of freedom. 

\section{Conclusions}

We have introduced the concept of geometric algebra
as the algebra of a vector space and all its subspaces. 
We have shown how conformal geometric algebra embeds and models
Euclidean geometry. Outer products of points (including the
point at infinity) model the elementary geometric objects
of points, point pairs, flat points, circles, lines, spheres,
planes and the embedded 3D space itself. 

The unified representation of affine Euclidean transformations (including
translations) by orthogonal transformations in the
conformal model allows the construction of
a geometric neuron, whose versor weights can learn these transformations
precisely. The transformations were illustrated in detail.




\subsubsection*{Acknowledgments} 

The author would like to thank K. Tachibana (COE FCS, Nagoya)
and S. Buchholz (Kiel).
He further wishes to thank God, the creator: 
\textit{Soli Deo Gloria},
and his very supportive family.


\begin{thebibliography}{99}

\bibitem{AH:book}
  A. Hirose,
  \textit{Complex-Val. NNs}, 
  Springer, Berlin, 2006.

\bibitem{BS:ESANN2000}
  S. Buchholz, G. Sommer, 
  \textit{Quat. Spin. MLP}, 
  Proc. ESANN 2000, 
  d-side pub., pp. 377--382, 2000

\bibitem{NM:Matsui}
  N. Matsui, et al,
  \textit{Quat. NN with geome. ops.},
  J. of Intel. \& Fuzzy Sys. \textbf{15} pp. 149-1164 (2004).

\bibitem{EH:MVDC}
  E. Hitzer,
  \textit{Multivector Diff. Calc.},
  Adv. Appl. Cliff. Alg. \textbf{12(2)} pp. 135-182 (2002).

\bibitem{DH:CtoG}
  D. Hestenes, G. Sobczyk, 
  \textit{Cliff. Alg. to Geom. Calc.}, 
  Kluwer, Dordrecht,   1999.

\bibitem{SB:thesis}
  S. Buchholz, 
  \textit{A Theory of Neur. Comp. with Cliff. Alg.}, 
  TR No. 0504, Univ. of Kiel, May 2005.
  
\bibitem{HLR:GS1}
  D. Hestenes, H. Li, A. Rockwood, 
  \textit{New Alg. Tools for Class. Geom.},
  in G. Sommer (ed.), Geom. Comp. with Cliff. Alg.,
  Springer, Berlin, 2001.

\bibitem{DFM:GACS}  
  L. Dorst, D. Fontijne, S. Mann,
  {\it Geom. Alg. for Comp. Sc.},
  Morgan Kaufmann Ser. in Comp. Graph., 
  San Francisco, 2007.

\bibitem{PA:ConfGr}
  P. Angles,
  \textit{Conf. Groups in Geom. and Spin Struct.}, 
  PMP, Birkhauser, Boston, 2007.
  
\bibitem{PL:Clifford}
   P. Lounesto,
   {\it Cliff. Alg. and Spinors}, 2nd ed.,
   CUP, Cambridge, 2006.

\bibitem{BHT:ICANN2007}
  S. Buchholz, K. Tachibana, E. Hitzer, 
  \textit{Optimal Learning Rates for Cliff. Neurons},
  in LNCS 4668, Springer, Berlin, 2007, pp. 864--873.
  
\bibitem{HBTY:WCAA}
  E. Hitzer, et al.
  \textit{Carrier method f. gen. eval. \& control of pose, 
  molec. conform., tracking, and the like}
  acc. Adv. in Appl. Cliff. Algs., 26 pp. (2008). 
  
\bibitem{BHT:CVN}
  S. Buchholz, E. Hitzer, K. Tachibana,
  \textit{Coordinate independent update formulas for versor
Clifford neurons}
  Proceedings of Joint Conference SCIS \& ISIS 2008, Nagoya, Japan, pp. 814--819, (2008).
  
\bibitem{CP:CLUCalc}
  C. Perwass, Visual Calculator CLUCalc, http://www.clucalc.info
  
\bibitem{HP:SGV}
  C. Perwass, E. Hitzer, Space Group Visualizer, http://www.spacegroup.info

\end{thebibliography}
\end{document}